\documentclass[]{openmoss}

\usepackage{helvet}

\usepackage{amsmath} 
\usepackage{natbib}
\usepackage{graphicx}
\usepackage{subcaption} 

\usepackage[toc,page,header]{appendix}
\usepackage[utf8]{inputenc} 
\usepackage[T1]{fontenc}    
\usepackage{hyperref}       
\usepackage{url}            
\usepackage{booktabs}       
\usepackage{lmodern}        
\usepackage{amsfonts}       
\usepackage{nicefrac}       
\usepackage{microtype}      
\usepackage{wrapfig}

\usepackage{titletoc}
\usepackage{minitoc}
\usepackage{algorithm}
\usepackage{algpseudocode}
\usepackage{amsmath}
\usepackage{amssymb}

\usepackage{array}
\usepackage{etoolbox}

\definecolor{lightblue}{RGB}{200, 230, 255}  
\definecolor{headerblue}{RGB}{150, 200, 255} 

\usepackage{pgfplots}
\usepackage{xcolor}
\usepackage{float}
\usepackage{comment}
\usepackage{multirow}
\usepackage{makecell}
\usepackage{siunitx}
\usepackage{tikz}
\usepackage{pgf-pie}
\usepackage[export]{adjustbox}

\usepackage{ragged2e}
\usepackage{tabularx}
\usepackage{caption}
\usepackage{enumitem}
\usepackage{pifont}
\usepackage[hang,flushmargin]{footmisc}

\usepackage{tcolorbox}
\tcbuselibrary{breakable}
\tcbuselibrary{skins}

\usepackage{tabularx}
\usepackage{listings}

\usepackage{threeparttable}
\usepackage{setspace}
\usepackage[scheme=plain]{ctex} 

\definecolor{MossCyan}{HTML}{82D9FF} 
\definecolor{MossBlue}{HTML}{82B1FF}

\definecolor{ForestGreen}{RGB}{34, 139, 34}
\definecolor{Red}{RGB}{255, 0, 0}

\definecolor{tickG}{rgb}{0.1, 0.588, 0.1}
\definecolor{crossR}{rgb}{0.588, 0.1, 0.1}

\definecolor{frenchblue}{rgb}{0.0, 0.45, 0.73}
\definecolor{babyblue}{rgb}{0.54, 0.81, 0.94}
\definecolor{classicrose}{rgb}{0.98, 0.8, 0.91}
\definecolor{beige}{rgb}{0.96, 0.96, 0.86}
\definecolor{forestgreen}{HTML}{2e7d43}

\definecolor{blue1}{HTML}{91BBE6}
\definecolor{blue2}{HTML}{3F90E0}
\definecolor{blue3}{HTML}{316FAD}

\definecolor{color1}{HTML}{FF9999}
\definecolor{color2}{HTML}{FF6666}
\definecolor{color3}{HTML}{FF3333}
\definecolor{color4}{HTML}{E60000}
\definecolor{color5}{HTML}{B30000}
\definecolor{color6}{HTML}{8CD98C}
\definecolor{color7}{HTML}{53c653}
\definecolor{color8}{HTML}{00B050}
\definecolor{color9}{HTML}{2d862d}
\definecolor{color10}{HTML}{206020}
\definecolor{color11}{HTML}{cca300}

\newtcolorbox{promptbox}[2][]{
    colback=white,
    coltext=black,
    arc=3mm,
    boxrule=0.5pt,
    colframe=black!60!white,
    title={#2},
    colbacktitle=black,
    coltitle=white,
    fonttitle=\bfseries,
    top=8pt,
    bottom=8pt,
    left=10pt,
    right=10pt,
    breakable,
    before upper={%
        \linespread{1}\selectfont
        \setlength{\parskip}{1ex plus 0.2ex minus 0.2ex}%
        \setlength{\parindent}{0pt}%
    },
    #1
}

\title{Advancing Omnimodal Embodied Agents from \\ Isolated Skills to Everyday Physical Autonomy}

\author{
Junhao Shi$^{1,2,*}$ \hspace{.3em}
Zezheng Huai$^{2,3,*}$\hspace{.3em}
Siyin Wang$^{1,2}$ \hspace{.1em}
Jia Chen$^{2}$\hspace{.1em}
Yubang Wang$^{2}$ \hspace{.1em}
\\
\textbf{
Zhaoye Fei$^{1}$ \hspace{.1em}
Hechang Chen$^{2,3}$ \hspace{.1em}
Jingjing Gong$^{2}$ \hspace{.2em}
Xipeng Qiu$^{1,2,\dagger}$ \hspace{.2em}
Yu-Gang Jiang$^{1,\dagger}$ 
}
\\
[1ex]
\texttt{24110240071@m.fudan.edu.cn, 253208540294@sii.edu.cn} \\
[1ex]
$^{1}$Fudan University   
$^{2}$Shanghai Innovation Institute
$^{3}$Jilin University
\\
}

\abstract{
Building persistent embodied agents in unstructured environments demands unified orchestration of heterogeneous tools spanning both cyber (APIs, IoT) and physical (manipulation, navigation) domains, coupled with autonomous recovery from physical failures that inevitably arise over extended operation. Existing systems treat these as separate problems: VLM-based planners lack a unified cyber-physical action space, agent frameworks accumulate unbounded context that degrades temporal coherence, and VLA policies execute open-loop without detecting their own failures. We argue that persistent autonomy requires not a monolithic model but a hierarchical asynchronous architecture with explicit separation of planning, memory, and verification. To this end, we present \textbf{OmniAct}, a framework integrating a multimodal semantic planner for skill routing across unified action spaces, an adaptive hierarchical memory with event-boundary-driven compression for sub-linear context growth, and an asynchronous visual preemption engine that closes the semantic loop during physical execution. Across 40 real-world long-horizon tasks on two robotic platforms coordinating four IoT devices, OmniAct achieves consistent improvements in end-to-end success across all complexity levels, maintains near-flat token consumption over under 100k+ accumulated interaction tokens, and elevates mid-scale open-weight models to proprietary-level performance.
}

\checkdata[Project Page]{\url{https://huaizz-shawen.github.io/OmniAct}}
\checkdata[GitHub Code]{\url{https://github.com/EmbodiedForge/RAS_interactivate_planner}}

\begin{document}
\maketitle
\begingroup
\renewcommand{\thefootnote}{\fnsymbol{footnote}}
\setcounter{footnote}{1}
\footnotetext{Equal Contribution. $\dagger$ Corresponding Authors.}
\endgroup




\section{Introduction}

The fundamental goal of embodied intelligence is to build persistent physical agents that operate reliably in unstructured real-world environments over extended temporal horizons. This requires the simultaneous integration of multimodal perception spanning speech, vision, and language, unified actuation across heterogeneous tools including IoT devices, Web APIs, robotic manipulators, and mobile platforms, as well as autonomous recovery from unforeseen physical failures. However, current embodied systems, whether end-to-end vision-language-action models or LLM-based planning frameworks, encounter fundamental limitations when deployed in such demanding real-world scenarios~\cite{vlasurvey,wamsurvey,vlasurveyrealworld,embodiedsurvey}.

Existing approaches fall short in three critical aspects.  First, current systems treat tool invocation (APIs, IoT) and physical manipulation as separate pipelines, yet realistic user interactions through natural speech often implicitly require coordination of multiple heterogeneous tools within a single task flow~\cite{saycan, codeaspolicy, innermonologue}. The absence of a unified cyber-physical action space forces these methods to rely on brittle hand-crafted interfaces between domains, severely limiting their capacity to handle naturally interleaved instructions. Second, agent frameworks that linearly concatenate interaction history accumulate context rapidly during long-term deployment, increasing memory overhead, losing critical historical information, and introducing semantic conflicts between long-term preferences and recent states~\cite{react, reflexion, voyager, memgpt}. Third, though current end-to-end VLA policies~\cite{rt-2, openvla, pi0, pi0.5, roboomni} have achieved notable progress, their inherently open-loop nature cannot support autonomous operation: undetected physical failures cascade into irreversible faults from which the system cannot recover~\cite{vlasurveyrealworld}.

\begin{figure}[t]
    \centering
    \includegraphics[width=0.85\linewidth]{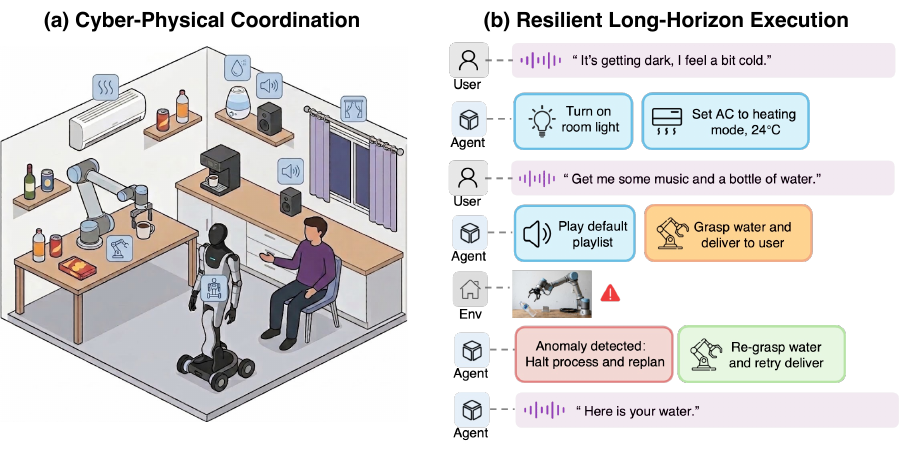}
    \vspace{-4pt}
    \caption{Two key challenges for real-world long-horizon embodied deployment. (a)~A single user command may require joint orchestration of IoT devices, robotic manipulators, mobile platforms, and web APIs. (b)~Over extended horizons, unexpected physical failures must be 
    autonomously detected and recovered from without human intervention.}
    \label{fig:teaser}
    \vspace{-5pt}
\end{figure}


As illustrated in Figure~\ref{fig:teaser}, achieving everyday physical autonomy faces two key challenges: (i)~coordinating heterogeneous cyber-physical tools within single task flows, and (ii)~autonomously recovering from physical failures over extended horizons. To address these challenges jointly, we observe that persistent real-world autonomy requires not a monolithic end-to-end model, but a hierarchical asynchronous architecture that explicitly decouples high-level planning, long-horizon state tracking, and real-time physical verification into specialized, cooperating modules. Consequently, we introduce \textbf{OmniAct}, an embodied framework achieving true ``omni-action'' by seamlessly bridging omnimodal perception (speech, vision, language) with a unified cyber-physical action space (IoT, Web APIs, robotic control). OmniAct comprises three tightly integrated modules: (1) a \textit{multimodal semantic planner} for structured skill-routing; (2) an \textit{adaptive hierarchical memory} utilizing event-boundary-driven compression to maintain sub-linear context growth; and (3) an \textit{asynchronous visual preemption engine} that periodically verifies physical execution and triggers immediate replanning upon detecting anomalies.

We evaluate \textbf{OmniAct} across 40 real-world long-horizon tasks on two structurally distinct platforms (UR5e manipulator and wheeled mobile robot) coordinating four household IoT devices. Experiments show that OmniAct substantially outperforms both open-loop and step-wise reasoning baselines, achieving higher end-to-end success and improved perturbation recovery(Sec.~\ref{sec:exp1}), reduced token consumption with near-flat context growth (Sec.~\ref{sec:exp2}), and strong cross-model generalizability (Sec.~\ref{sec:exp4}). 

Our contributions are threefold:

\begin{enumerate}
    \item We introduce \textbf{OmniAct}, a multimodal embodied framework that unifies discrete cyber tools (APIs, IoT) and continuous physical control into a single event-driven loop with asynchronous visual preemption, enabling closed-loop persistent deployment across diverse robotic platforms without force sensing.
    \item We design an adaptive hierarchical memory that condenses long-horizon interaction history into timestamped semantic cues through event-boundary-driven compression and integrates explicit physical reflection, enabling the agent to maintain bounded context growth while autonomously recovering from failures and adapting to dynamic environmental shifts.
    \item We conduct extensive real-world empirical validation across diverse physical environments and robotic platforms. Compared to reactive policies and LLM-based planning baselines, our framework significantly improves end-to-end success rates in failure-prone scenarios while substantially reducing token consumption and inference latency.
\end{enumerate}

\section{Related Work}
\subsection{Embodied Agents with Foundation Models}

The integration of foundation models into embodied systems has progressed rapidly. Early work leverages large language models as high-level planners that decompose natural language instructions into executable skill sequences, either through affordance grounding~\cite{saycan}, code generation~\cite{code_as_policies, progprompt}, or iterative replanning with textual feedback~\cite{inner_monologue}. Subsequent agent architectures extend these planners with autonomous reasoning capabilities, incorporating chain-of-thought tool invocation~\cite{react}, verbal self-reflection for trial improvement~\cite{reflexion}, and persistent skill accumulation for open-ended exploration~\cite{voyager}. More recent efforts further integrate multimodal perception into the agentic loop: PaLM-E~\cite{palme} grounds planning in visual and sensor observations, WAP~\cite{WAP} advances planning toward closed-loop continuous setting. RoboBrain~\cite{robobrain,robobrain2} combines manipulation planning with affordance reasoning, and hierarchical frameworks such as RT-H~\cite{rth} bridge high-level language to low-level motor execution through structured sub-goal decomposition.

Despite this progress, existing approaches operate within either the cyber or physical domain without a unified action space,  mostly focus on short interactions without addressing long-horizon context accumulation toward real-world autonomy. OmniAct addresses these gaps by unifying cyber tools and physical control into a single skill-routing space, introducing event-boundary-driven hierarchical memory for bounded context growth, and employing asynchronous visual preemption for semantic-level closed-loop verification.

\subsection{Vision-Language-Action Models}

Vision-Language-Action (VLA) models directly map visual observations and language instructions to low-level control signals. Early architectures explored various fusion strategies including feature modulation~\cite{bcz, rt1}, cross-attention~\cite{vima, rt-2}, and token concatenation~\cite{gato}. Subsequent large-scale pre-training on diverse robotic datasets significantly improved model capabilities~\cite{openvla, octo}. Concurrently, advances in action representation such as autoregressive tokenization~\cite{fast} and diffusion-based generation~\cite{pi0, rdt1b, pi0.5}, together with richer input modalities including 3D geometry~\cite{3dvla} and tactile feedback~\cite{tavla, tactilevla}, have further enhanced generalization across environments and task horizons. Most recently, RoboOmni~\cite{roboomni} extends VLA to accept speech input and enable proactive interaction, moving closer to natural human-robot communication.

Nevertheless, current VLA models remain limited in cross-embodiment transfer, multi-task versatility, and zero-shot generalization to novel scenarios. More critically, they lack inherent reasoning capabilities and cannot autonomously detect or correct their own execution failures. Omni does not seek to replace VLA models but instead incorporates them as low-level executors within a unified orchestration framework, compensating for these limitations through asynchronous visual verification for failure detection and hierarchical memory for long-horizon state coherence.

\section{Methodology: The OmniAcr Framework}

\begin{figure}[t]
    \centering
    \includegraphics[width=1\linewidth]{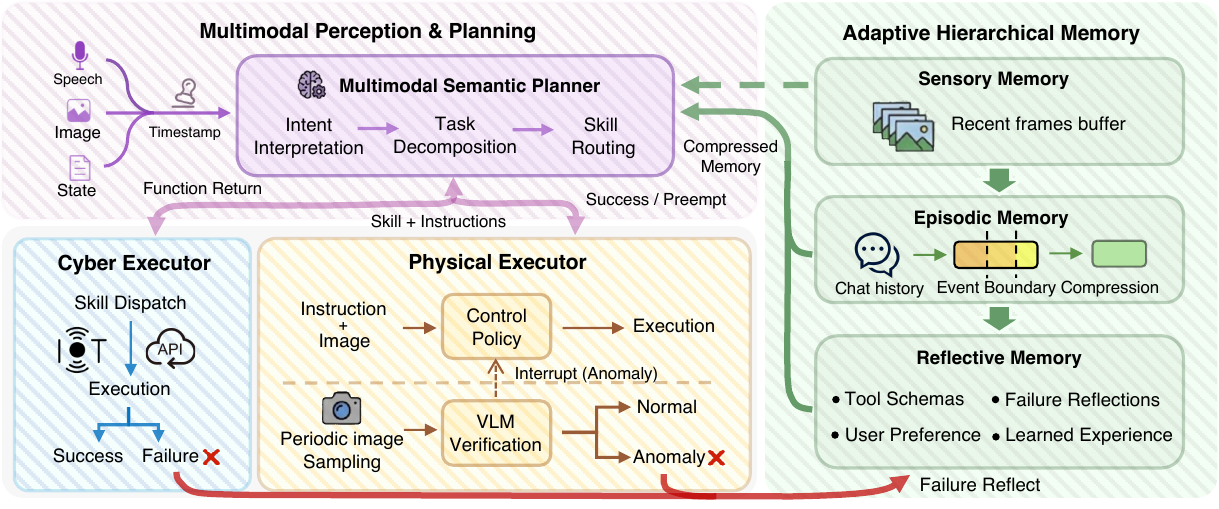}
    \caption{\textbf{Overview of the OmniAct framework.} The system comprises three modules: (1) \textit{Multimodal Semantic Planner} fuses speech, image, and state inputs for intent interpretation, task decomposition, and skill routing; (2) \textit{Closed-Loop Execution Engine} with asynchronous visual preemption that periodically verifies physical progress via a VLM and triggers interruption upon anomaly; and (3) \textit{Adaptive Hierarchical Memory} (sensory, episodic, reflective) employing event-boundary compression for bounded context growth and storing failure reflections for cumulative adaptation.}
    \label{fig:framework}
\end{figure}

\subsection{Problem Formulation}

A key challenge in building persistent embodied agents is to enable a single planning system to coordinate diverse executors spanning both cyber (APIs, IoT devices) and physical (robotic manipulation, navigation) domains under partial observability. We formulate this as a decision process over a unified cyber-physical action space.

At each time step $t$, the system receives a multimodal observation $o_t = (v_t, s_t)$, where $v_t$ is a visual frame and $s_t$ denotes discrete state information from sensors or tool APIs. The multimodal context $c_t$ integrates $o_t$ with the current user intent (e.g., transcribed speech) and system state. Since the system must dispatch to fundamentally different executors ranging from stateless API calls to continuous robotic control, a single low-level action representation is insufficient. We therefore define a skill-routing action space: each action is a tuple $a_t = (\tau_t, g_t)$, where $\tau_t \in \mathcal{T}$ selects the target skill and $g_t$ specifies skill-specific conditioning parameters. The historical trajectory $\mathcal{H}_t$, retrieved from hierarchical memory (Sec.~\ref{sec:memory}), provides compressed prior experience. Given the planning model $\pi$, the high-level planning policy is:
\begin{equation}
    a_t \sim \pi(\cdot \mid c_t, \mathcal{H}_t, \mathcal{T}).
\end{equation}

\subsection{Multimodal Semantic Planner}
The semantic planner instantiates model $\pi$ and is responsible for task decomposition, skill selection, and sub-goal generation. Given context $c_t$ and memory $\mathcal{H}_t$, the planner performs three functions: (1) interpreting potentially ambiguous user intent by grounding speech and visual observations against the current tool state; (2) decomposing long-horizon instructions into executable sub-goals, selecting appropriate skills $\tau_t$ from inventory $\mathcal{T}$ based on task requirements and environmental constraints; and (3) leveraging historical experience from $\mathcal{H}_t$ to avoid previously observed failures and respect long-term user preferences. The planner outputs a structured JSON specification supporting parallel dispatch of multiple skill calls within a single step. Details is provided in the Appendix.

\subsection{Closed-Loop Execution with Asynchronous Visual Preemption}
A fundamental limitation of VLA policies is their open-loop nature: once committed, the executor cannot perceive whether the physical world has deviated from expectations, and a single undetected failure propagates into cascading faults. OmniAct addresses this through an asynchronous visual preemption mechanism that maintains a low-frequency semantic closed-loop concurrent with physical execution.

The execution engine dispatches actions along two parallel streams. Discrete cyber commands (IoT control, API calls) are verified through deterministic return codes. Continuous physical commands are forwarded to the control policy, for which no such verification signal exists. To compensate, an asynchronous visual monitor periodically samples frames and submits them to a VLM for semantic-level assessment of execution progress and physical safety. If anomaly is detected, the system immediately halts the VLA executor, packages the failure context, and returns control to the planner for replanning.

Unlike geometric constraints in model-predictive control that verify joint limits or collision distances, this mechanism reasons about high-level task semantics, whether the target object remains grasped, whether the robot is progressing toward the goal—capturing failure modes invisible to low-level controllers. Upon preemption or failures from the cyber executor, the failure episode simultaneously triggers a reflection process that stores causal analysis and corrective strategies in long-term memory (Sec.~\ref{sec:memory}) to prevent recurrence.

\subsection{Adaptive Hierarchical Memory}
\label{sec:memory}
A persistent agent must maintain awareness of its full interaction history, yet naively accumulating raw observations causes linear context growth that quickly exceeds model capacity. Our key observation is that physical interaction naturally segments into discrete semantic events, and compression should operate at these event boundaries rather than at fixed intervals or uniform turn-level granularity. We design a three-level memory hierarchy with increasing abstraction:

\textbf{\textit{Sensory Memory.}} A lightweight FIFO queue maintaining recent visual frames for the control policy and visual monitor, decoupled from the planning context.

\textbf{\textit{Episodic Memory.}} This level manages interaction history through event-boundary-driven compression. Each record is annotated with absolute timestamps for temporal ordering. Upon detecting a semantic boundary (e.g., sub-task completion, state change), the raw observation sequence is compressed into a compact keyframe summary retaining semantic content, state transitions, and action outcomes while discarding procedural redundancy. Context thus grows proportionally to distinct semantic events rather than raw turns, yielding sub-linear scaling over extended deployment.

\textbf{\textit{Reflective Memory.}} This level stores both static priors (tool schemas, skill inventory) and dynamically accumulated experience. When execution failure or visual preemption is triggered, the system generates a structured record of root cause and corrective strategy. During subsequent planning, relevant reflections are retrieved and injected into context, transforming isolated error recovery into cumulative adaptation over the agent's operational lifetime.

\section{Experiments}

We structure our evaluation around three axes: cross-domain orchestration across heterogeneous hardware (Sec.~\ref{sec:exp1}), memory effectiveness and scalability under long-horizon deployment (Sec.~\ref{sec:exp2}), and generalizability across foundation models of varying scale (Sec.~\ref{sec:exp4}).

\subsection{Experimental Setup}

\paragraph{\textbf{Platforms and Tools.}} We evaluate OmniAct on two robotic platforms with distinct morphologies: a UR5e 6-DoF manipulator for precision tabletop manipulation and a Keenon wheeled mobile robot for indoor navigation. The system interfaces with 4 household IoT devices (lighting, climate control, appliances) and 12 virtual tool APIs (web search, checklist, delivery status, etc.), forming a heterogeneous cyber-physical environment representative of realistic domestic deployment scenarios.

\paragraph{\textbf{Task Suite.}} To reflect the multimodal, cross-domain nature of realistic human-robot interaction, we construct a benchmark of 40 real-world tasks divided into two categories by physical modality: \textit{manipulation} (20 tasks, deployed on UR5e) and \textit{navigation} (20 tasks, deployed on Keenon mobile robot) , each grouped into three complexity levels by modality involvement and coordination depth:

\textit{Level 1 (Physical Only, 20\%)}: Tasks requiring only low-level robotic control (e.g., pick-and-place).

\textit{Level 2 (Cyber-Physical, 40\%)}: Tasks additionally requiring IoT device coordination or web API querie (e.g., activating an appliance before manipulation).

\textit{Level 3 (Full Multimodal, 40\%)}: Tasks involving speech interaction, web API queries, IoT control, and physical execution in an integrated flow.

This stratification exposes how gracefully each method scales with cross-modal coordination depth while ensuring fair comparison on subsets accessible to all baselines. Each task is repeated 3 times with success adjudicated by human evaluation. The complete task list is provided in the appendix.

\paragraph{\textbf{Baselines.}} We compare OmniAct against four methods: \textit{Direct Policy}, our self-trained VLA / navigation policies without planning or memory (L1 difficulty only); \textit{SayCan}~\cite{saycan}, affordance-grounded scoring with LLM planning; \textit{Code as Policy}~\cite{code_as_policies}, one-shot code generation planning for full task plans; and \textit{RoboBrain2}~\cite{robobrain2}, a 32B embodied vision-language model that unifies perception, reasoning, and planning, surpassing proprietary models on multiple embodied benchmarks~\cite{robobrain2,egoplan2}. All methods share identical tool interfaces. Except RoboBrain2, which uses its own backbone, all employ Gemini-3.1-Pro to isolate architectural differences.


\begin{figure*}[t]
\centering
\begin{subfigure}{\textwidth}
    \centering
    \includegraphics[width=0.85\textwidth]{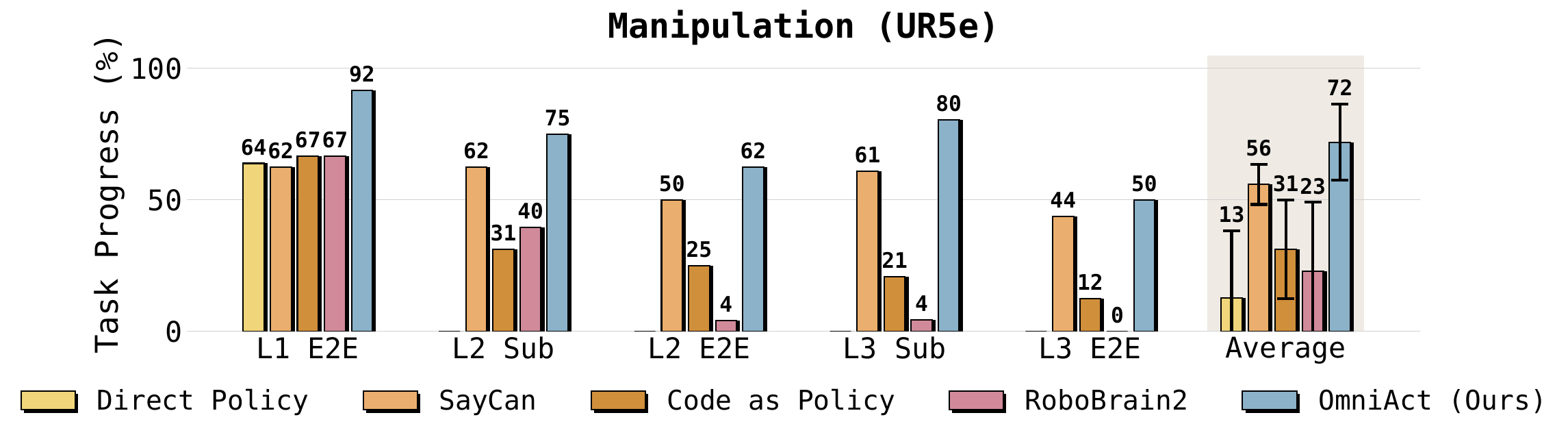}
    \label{fig:manipulation}
\end{subfigure}
\vspace{0.3cm}
\begin{subfigure}{\textwidth}
    \centering
    \includegraphics[width=0.85\textwidth]{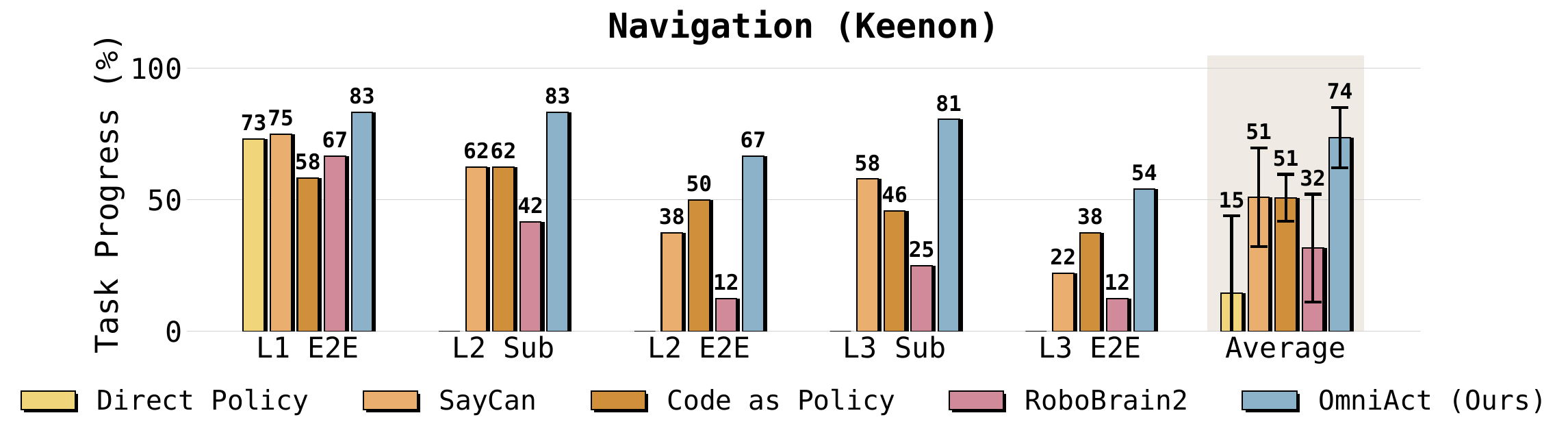}
    \label{fig:navigation}
\end{subfigure}
\caption{End-to-end and sub-task success rates (\%) on long-horizon cyber-physical tasks across three complexity levels. Performance is evaluated on two structurally distinct real-world platforms: a UR5e manipulator (top) and a Keenon mobile robot (bottom). While baseline methods exhibit severe performance degradation as cross-modal coordination depth increases from L1 to L3, OmniAct demonstrates a significantly more graceful decline, substantially outperforming prior approaches on both partial progress (Sub) and strict task completion (E2E). Error bars in the rightmost column denote standard deviation. Empty bars indicate that the baseline lacks the architectural capability for that specific complexity level; these are counted as zero when computing the overall average.}
\label{fig:main_results}
\end{figure*}

\subsection{Long-Horizon Cyber-Physical Orchestration}
\label{sec:exp1}
To disentangle partial progress from true task completion, we report two complementary metrics: \textit{Sub-task Success Rate (Sub)}, the average fraction of correctly completed sub-tasks, and \textit{End-to-End Success Rate (E2E)}, a strict binary measure crediting only full task completion. The gap between them quantifies how severely single-stage errors cascade through the pipeline, a critical diagnostic for long-horizon autonomy where an unrecovered failure can invalidate all subsequent stages.


Figure~\ref{fig:main_results} reveals a consistent trend: as task complexity increases from L1 to L3, all baselines suffer steep performance degradation, while OmniAct maintains comparatively graceful decline. 
On the most demanding L3 tasks, OmniAct achieves 50.0\% and 54.2\% E2E success on manipulation and navigation respectively, significantly exceeding the best baseline (SayCan: 43.8\% on manipulation; Code as Policy: 37.5\% on navigation).
Furthermore, OmniAct achieves the highest Sub-task success across all complexity levels, with L3 Sub reaching 80.5\% on manipulation and 80.6\% on navigation, compared to the best baseline scores of 60.9\% and 57.9\% respectively.
This directly validates the contribution of asynchronous visual preemption and reflective memory: when a VLA execution fails mid-task (e.g., grasp slippage), baselines proceed obliviously and corrupt all downstream stages, whereas OmniAct detects the deviation, halts execution, and replans from the current state, converting catastrophic failures into recoverable delays.

An instructive case is RoboBrain2, which despite achieving state-of-the-art on established embodied planning benchmarks, obtains a surprisingly low score. Its failures concentrate on generating well-formed API calls and maintaining state across interleaved cyber-physical sub-tasks, likely compounded by the absence of IoT-domain training data, reinforcing that cross-modal tool coordination is a fundamentally different competency from spatial scene understanding and demands an explicit orchestration layer.

Finally, the consistency of gains across both the fixed manipulator and the mobile robot deserves emphasis. Despite their fundamentally different kinematic structures, workspace scales, and task semantics, OmniAct achieves comparable relative improvements on both platforms without any platform-specific architectural modification or additional tuning. This confirms that the framework's core contributions—unified skill routing, hierarchical memory, and visual preemption—operate at an abstraction above embodiment specifics and transfer directly to new hardware configurations.

\subsection{Adaptive Hierarchical Memory}
\label{sec:exp2}

\paragraph{\textbf{Temporal Conflict Resolution.}}
To evaluate whether OmniAct's memory preserves long-term coherence under extended deployment, we construct three scenarios (tool use, home assistant, robotic manipulation), each containing 10 samples exceeding 10 hours of interaction with 40k+ accumulated tokens. Temporal preference conflicts are embedded throughout: early turns establish persistent constraints (e.g., no sugar'') while later requests introduce seemingly conflicting needs (e.g., something flavored''), requiring the agent to satisfy both simultaneously. We ablate against four variants: \textit{Raw History} concatenates all past turns verbatim; \textit{Sliding Window} retains only the most recent 10 turns; \textit{Turn Summary} compresses each turn independently and concatenates linearly; \textit{Episodic Only} applies event-boundary compression without reflective memory.

\begin{wraptable}{r}{0.6\textwidth}
\centering
\small
\vspace{-5mm}
\caption{Constraint Satisfaction Rate (\%) and context compression ratio across three long-horizon scenarios. Compression is relative to raw history length.}
\label{tab:memory}
\resizebox{0.6\textwidth}{!}{
\begin{tabular}{lcccc}
\toprule
Method & Tool Use & Home Asst. & Manip. & Compress. \\
\midrule
Raw History & 66.67\% & 46.67\% & 66.67\% & 100.0\% \\
Sliding Window & 13.33\% & 6.67\% & 6.67\% &16.3\%\\
Turn Summary & 56.67\% & 50.0\% & 63.33\% & 27.4\%\\
Episodic Only & 56.67\% & 46.67\% & 66.67\% & 21.2\% \\
\midrule
OmniAct (Full) & \textbf{86.67\%} & \textbf{66.67\%} & \textbf{80.0\%} & 24.5\% \\
\bottomrule
\end{tabular}
}
\vspace{-10pt}
\end{wraptable}

As shown in Table~\ref{tab:memory}, Sliding Window achieves the most aggressive compression (16.3\%) but catastrophically discards early-established constraints, collapsing to near-zero satisfaction across all scenarios—demonstrating that recency alone is entirely insufficient for preference-aware long-horizon agents. Raw History retains all information yet performs only moderately (46.67\%--66.67\%): the sheer volume of 40k+ tokens introduces positional dilution that makes relevant constraints difficult to locate, and semantic conflicts between temporally distant preferences often mislead the planner. Turn Summary and Episodic Only both achieve meaningful compression while maintaining reasonable accuracy, but neither consistently surpasses Raw History—indicating that compression without explicit long-term knowledge retrieval trades one failure mode (dilution) for another (information loss). OmniAct's full system adds reflective memory atop episodic compression, boosting satisfaction by 13--20 points over the best ablation variant. The gain stems from the reflective layer explicitly surfacing long-term constraints and failure-derived corrective strategies at planning time, rather than relying on the planner to implicitly extract them from compressed episode sequences.

\paragraph{\textbf{Long-Horizon Scalability.}}

\begin{wrapfigure}{r}{0.6\textwidth}
\centering
\vspace{-1mm}
\includegraphics[width=0.6\textwidth]{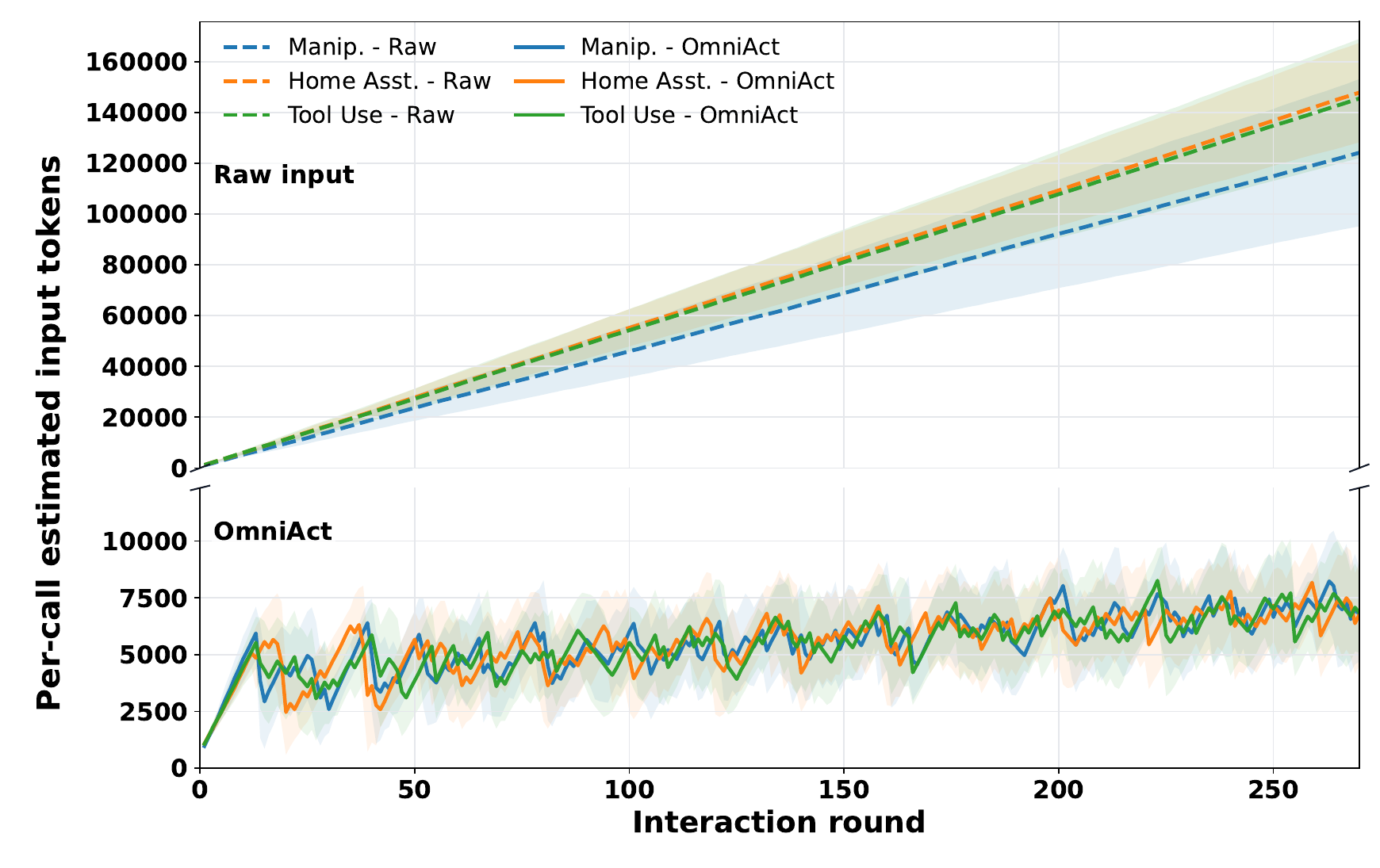}
\caption{Per-call estimated input tokens over 270 interaction rounds across three scenarios. \textit{Top}: raw history concatenation grows linearly, exceeding 160k tokens. \textit{Bottom}: OmniAct's hierarchical memory remains bounded within 2.5k--10k tokens regardless of accumulated history length.}
\label{fig:token_growth}
\vspace{-10pt}
\end{wrapfigure}

To stress-test scalability under continuous deployment, we extend the interaction traces to 100k+ tokens (approximately 270 rounds) and track per-call input token consumption over time. As shown in Figure~\ref{fig:token_growth}, raw history replay exhibits strictly linear growth, exceeding 120k tokens by round 260 and approaching the context window limits of most foundation models—at which point either truncation or degraded generation quality becomes inevitable. In contrast, OmniAct's event-boundary-driven compression maintains per-call token usage within a narrow 4k--8k band that plateaus early and fluctuates only with local episode complexity rather than accumulated history length. This near-flat scaling profile fundamentally decouples deployment duration from inference cost, enabling persistent operation over arbitrarily long horizons without context overflow or progressive quality degradation.


\subsection{Cross-Model Generalizability}
\label{sec:exp4}

\begin{wraptable}{r}{0.6\textwidth}
\centering
\small
\vspace{-5mm}
\caption{Constraint Satisfaction Rate (\%) with and without OmniAct across foundation model backbones. $\Delta$ represents absolute improvement.}
\label{tab:backbone}
\resizebox{0.6\textwidth}{!}{
\begin{tabular}{lccc}
\toprule
Backbone & Raw & OmniAct & $\Delta$ \\
\midrule
Gemini-3.1-Pro-Preview & 76.7 & 80.0 & +3.3 \\
Gemini-3-Flash-Preview & 70.0 & 76.7 & +6.7 \\
Qwen3.6-Plus & 63.3 & 80.0 & +16.7 \\
Qwen3.6-Flash & 56.7 & 73.3 & +16.6 \\
Qwen3-VL-30B-A3B & 50.0 & 80.0 & +30.0 \\
Qwen3-VL-8B & 30.0 & 36.7 & +6.7 \\
\bottomrule
\end{tabular}
}
\end{wraptable}

To determine whether OmniAct's gains stem from architectural design rather than backbone strength, we replace the planner across six models spanning proprietary (Gemini-3.1-Pro, Gemini-3-Flash, Qwen3.6-Plus/Flash) and open-weight (Qwen3-VL-30B-A3B, Qwen3-VL-8B) families, using the same long-horizon setting as Sec.~\ref{sec:exp2}.

As shown in Table~\ref{tab:backbone}, OmniAct improves all backbones with sufficient baseline capability. Mid-tier models exhibit the most substantial gains: Qwen3-VL-30B-A3B improves from 50.0\% to 80.0\% (+30.0), reaching parity with Gemini-3.1-Pro's vanilla performance and demonstrating that OmniAct's structured orchestration can close the gap between open-weight and proprietary systems without additional training. At the upper end, even Gemini-3.1-Pro benefits (+3.3), indicating that hierarchical memory and asynchronous verification address systematic failure modes—such as long-horizon preference drift and undetected execution errors—that persist regardless of model scale. At the lower end, Qwen3-VL-8B shows only marginal improvement (+6.7), as its limited capacity to reliably produce well-formed structured outputs constrains the framework's ability to orchestrate multi-step tool invocations—establishing an approximate lower bound on backbone capability required for effective integration.

\section{Conclusions}
\label{sec:conclusion}
We presented OmniAct, a multimodal embodied framework that addresses persistent real-world autonomy by decomposing planning, memory, and verification into three cooperating modules: a unified cyber-physical semantic planner, an event-boundary-driven hierarchical memory for bounded context growth, and an asynchronous visual preemption engine for closed-loop failure recovery. Real-world experiments across diverse robotic platforms and backbone models demonstrate consistent improvements over both monolithic VLA and LLM-based baselines, with near-flat token consumption over extended horizons. Notably, OmniAct elevates mid-scale open-weight models to proprietary-level performance without additional training, underscoring that architectural clarity and structured orchestration complement model capacity rather than competing with it. These results suggest that explicit separation of responsibilities, rather than monolithic model scaling, offers a more viable and resource-efficient path toward persistent embodied deployment in unstructured real-world environments.

\clearpage
\bibliographystyle{unsrtnat} 
\bibliography{main}

\clearpage
\beginappendix

\startcontents[app]
\begingroup
  \renewcommand{\contentsname}{Appendix Contents}
  \section*{\contentsname}
  \printcontents[app]{}{1}{}
\endgroup
\newpage



\newtcolorbox{mossbox}[1]{
    enhanced,                  
    colback=white,             
    colframe=MossBlue,         
    colbacktitle=MossBlue,    
    coltitle=black,           
    title style={left color=MossCyan, right color=MossBlue},
    title=\textbf{#1},         
    fontupper=\ttfamily,               
    boxrule=0.8pt,      
    left=2mm, right=2mm, top=2mm, bottom=2mm 
}

\newcolumntype{O}[1]{>{\raggedright\arraybackslash}p{#1}}

\tcbset{
  omniBox/.style={
    breakable,
    colback=gray!4,
    colframe=gray!45,
    coltitle=black,
    fonttitle=\bfseries,
    boxrule=0.4pt,
    arc=1mm,
    left=1mm,
    right=1mm,
    top=1mm,
    bottom=1mm
  },
  omniPrompt/.style={
    omniBox,
    listing only,
    listing options={
      basicstyle=\ttfamily\footnotesize,
      breaklines=true,
      columns=fullflexible,
      keepspaces=true,
      showstringspaces=false
    }
  }
}




\section{Limitations}
Despite the demonstrated improvements, OmniAct's adaptive capacity is fundamentally asymmetric: the high-level planner can leverage reflective memory to replan around failures, but cannot improve the frozen downstream VLA policies themselves—when failures stem from inherent motor limitations, recovery is bounded by the existing skill repertoire. Additionally, the asynchronous visual monitor operates at VLM inference frequency rather than control frequency, leaving a latency window during which erroneous actions may continue unchecked, posing residual safety risks in scenarios involving fragile objects or human-proximate operation.

\section{Real-World Experimental Setup}

\begin{figure}[ht]
\centering
\begin{subfigure}[t]{0.48\linewidth}
\centering
\IfFileExists{pic/ur5e_setup.png}{
  \includegraphics[height=0.23\textheight,keepaspectratio]{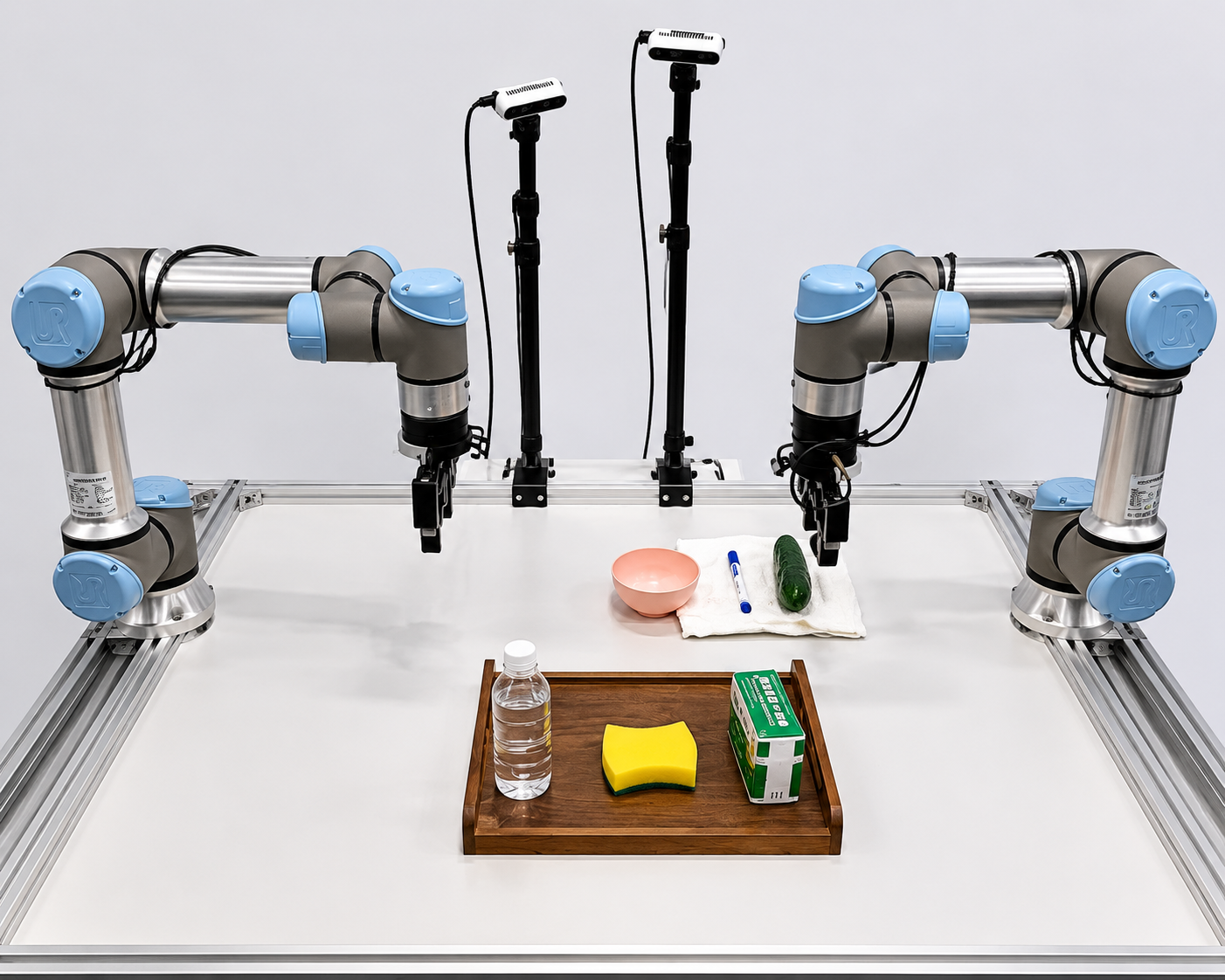}
}{
  \fbox{\begin{minipage}[c][0.23\textheight][c]{0.95\linewidth}
  \centering Placeholder for UR5e setup image.\\
  Replace \texttt{pic/ur5e\_setup.png}.
  \end{minipage}}
}
\caption{UR5e tabletop workspace with dual third-person and one
wrist-mounted RGB camera.}
\label{fig:setup-ur5e}
\end{subfigure}
\hfill
\begin{subfigure}[t]{0.48\linewidth}
\centering
\IfFileExists{pic/keenon_setup.png}{
  \includegraphics[height=0.23\textheight,keepaspectratio]{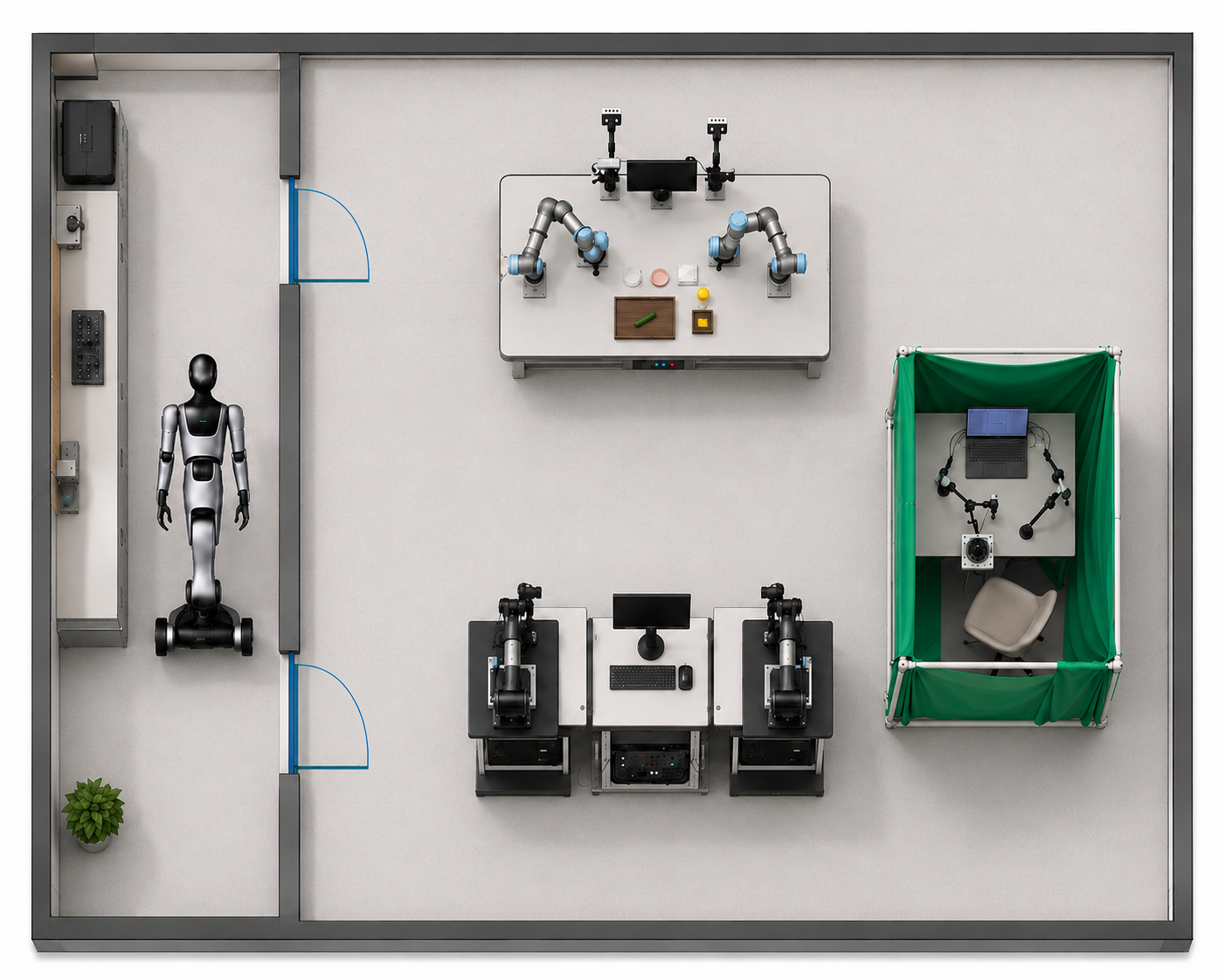}
}{
  \fbox{\begin{minipage}[c][0.23\textheight][c]{0.95\linewidth}
  \centering Placeholder for Keenon environment image.\\
  Replace \texttt{pic/keenon\_setup.png}.
  \end{minipage}}
}
\caption{Multi-room studio environment with the UR5e station, a
WidowX area, and a curtained side zone.}
\label{fig:setup-keenon}
\end{subfigure}
\caption{Real-world experimental scenes used in the long-horizon
cyber-physical benchmark. (a) The UR5e workspace serves as the
manipulation testbed for pick-and-place over everyday objects
(bottle, sponge, bowl, etc.). (b) The Keenon mobile robot navigates
between semantic locations (entrance, corner area, manipulation
station) and coordinates with the UR5e arm, enabling cross-platform
tasks that combine navigation, IoT control, and embodied manipulation
within a single execution loop.}
\label{fig:real-world-setup}
\end{figure}

OmniAct is evaluated on two structurally different real-world
platforms: a UR5e 6-DoF arm for tabletop manipulation, and a
Keenon wheeled mobile robot for indoor navigation. Both platforms
share the same high-level planner interface and differ only in the
exposed physical skills. The system also interfaces with four
household IoT devices (smart light, air conditioner, ambient audio
player, and a smart-home mode controller) and twelve API-style
virtual tools (web search, weather query, checklist retrieval, etc.).
Figures~\ref{fig:setup-ur5e} and~\ref{fig:setup-keenon} show the
two physical environments.


\section{Action Interface and Skill Implementations}
The high-level planner emits only structured skill calls; the actual
realization of each skill is delegated to backend executors.
Manipulation skill \texttt{pick\_and\_place} is implemented by a
self-trained $\pi_0$-style VLA policy fine-tuned on in-house
teleoperation data covering the supported object and location set.
Navigation skill \texttt{navigate\_to} is realized through a
pre-built 2D semantic map of the indoor environment with classical
waypoint following. Cyber skills (\texttt{control\_*},
\texttt{query\_*}, \texttt{web\_search}, \texttt{set\_home\_mode})
are deterministic JSON-returning HTTP calls.
Tables~\ref{tab:ur5e-actions} and~\ref{tab:keenon-actions} list the
full action interfaces exposed to the planner.

\begin{table}[t]
\centering
\small
\caption{UR5e manipulation action interface exposed to the semantic planner.}
\label{tab:ur5e-actions}
\renewcommand{\arraystretch}{1.55}
\setlength{\tabcolsep}{4pt}
\begin{tabular}{O{0.14\linewidth}O{0.19\linewidth}O{0.27\linewidth}O{0.31\linewidth}}
\hline
Action type & Action name & Parameters & Use \\
\hline
\texttt{talk} & \texttt{speak} & \texttt{message} & Final user-facing confirmation or essential status/error report. \\
\texttt{tool} & \texttt{store\_memory} & \texttt{content}, \texttt{scope}, \texttt{category} & Store durable user preference or operational facts only when explicitly relevant. \\
\texttt{tool} & \texttt{control\_light} & \texttt{action}, \texttt{device}, \texttt{brightness} & Turn lights on/off or set brightness. \\
\texttt{tool} & \texttt{play\_audio} & \texttt{action}, \texttt{audio\_type}, \texttt{track}, \texttt{volume} & Play or stop ambient audio. \\
\texttt{tool} & \texttt{web\_search} & \texttt{url}, \texttt{query} & Retrieve external information for multimodal tasks. \\
\texttt{tool} & \texttt{set\_home\_mode} & \texttt{device}, \texttt{option\_id}, \texttt{option\_text} & Select a predefined smart-device mode. \\
\texttt{act} & \texttt{pick\_and\_place} & \texttt{item\_name}, \texttt{source}, \texttt{target} & Move a supported visible object in the UR5e workspace. \\
\texttt{sense} & \texttt{get\_observation} & \texttt{\{\}} & Request a fresh observation when current visual evidence is insufficient. \\
\hline
\end{tabular}
\end{table}

\begin{table}[t]
\centering
\small
\caption{Keenon navigation and home-assistant action interface.}
\label{tab:keenon-actions}
\renewcommand{\arraystretch}{1.55}
\setlength{\tabcolsep}{4pt}
\begin{tabular}{O{0.14\linewidth}O{0.22\linewidth}O{0.27\linewidth}O{0.28\linewidth}}
\hline
Action type & Action name & Required parameters & Use \\
\hline
\texttt{talk} & \texttt{speak} & \texttt{message} & Final report, clarification, or completion notice. \\
\texttt{talk} & \texttt{send\_agent\_message} & \texttt{message}, \texttt{recipient} & Remote robot message when explicitly required. \\
\texttt{tool} & \texttt{control\_AC} & \texttt{action}, \texttt{temperature}, \texttt{fan\_speed} & Home air-conditioner control. \\
\texttt{tool} & \texttt{control\_light} & \texttt{device}, \texttt{action}, \texttt{brightness} & Lighting control. \\
\texttt{tool} & \texttt{set\_home\_mode} & \texttt{device}, \texttt{option\_id}, \texttt{option\_text} & Smart-home mode selection. \\
\texttt{tool} & \texttt{query\_weather\_api} & \texttt{location}, \texttt{query} & Weather or temperature query. \\
\texttt{act} & \texttt{navigate\_to} & \texttt{target\_location}, \texttt{with\_item} & Navigate to a home location; \texttt{with\_item} is set to \texttt{none}. \\
\texttt{sense} & \texttt{get\_observation} & \texttt{target} & Inspect a specified area after navigation. \\
\hline
\end{tabular}
\end{table}

\section{Task Suite Construction}
The benchmark contains 40 long-horizon real-world tasks, evenly split
between UR5e manipulation and Keenon navigation. For each platform we
design 4 Level-1 tasks (physical only), 8 Level-2 tasks
(cyber-physical), and 8 Level-3 tasks (full multimodal orchestration
with speech, web/API, IoT, and physical execution).
Tasks are designed to cover diverse object categories, IoT device
types, and API endpoints, avoiding repeated templates.
Each task is repeated three times under varied initial conditions and
adjudicated by a human evaluator using a binary end-to-end success
criterion. Table~\ref{tab:task-examples} lists representative requests
for each level; the complete task list will be released with the code.

\begin{itemize}
    \item \textbf{Level 1: Physical only.} Tasks require only physical robotic
    execution and visual grounding, such as object movement or visual
    inspection.
    \item \textbf{Level 2: Cyber-physical.} Tasks require physical execution
    together with IoT or API-style cyber actions.
    \item \textbf{Level 3: Full multimodal orchestration.} Tasks combine speech
    interaction, web/API access, IoT control, physical execution, and final
    natural-language reporting.
\end{itemize}

\begin{table}[t]
\centering
\small
\caption{Representative task requests for the three complexity levels.}
\label{tab:task-examples}
\renewcommand{\arraystretch}{1.55}
\setlength{\tabcolsep}{4pt}
\begin{tabular}{O{0.12\linewidth}O{0.17\linewidth}O{0.62\linewidth}}
\hline
Level & Modalities & Example request \\
\hline
L1 & VLA + vision & Pick up the pillbox from the tray and put it into the basket. \\
L2 & VLA + IoT + vision & Set the water heater's target temperature to $55^\circ$C. \\
L3 & VLA + IoT + web/API + speech & Check the simple steps for organizing the dishes, turn on the light to 80\% brightness, put the bowls into the basket, and report the result. \\
\hline
L1 & Navigation + vision & Go to the entrance area and check whether the umbrella is still there. \\
L2 & Navigation + IoT + vision & Turn up the brightness of the living-room light, then go to the living room and check whether the remote control is on the sofa. \\
L3 & Navigation + IoT + API + speech & Check whether it is going to rain today, turn on the entrance light, go to the entrance area and check whether the umbrella is still there, then report the result to me. \\
\hline
\end{tabular}
\end{table}

\section{Long-Horizon Context Construction}
\label{sec:longhorizon-construction}

In real deployment, task-relevant interactions are diluted by
environmental status checks, ambient logs, and unrelated household
exchanges. We mimic this regime by embedding a small set of
\emph{seed episodes} carrying task-relevant constraints (e.g.,
dietary preferences, accessibility needs, prior failure lessons)
within a much larger volume of \emph{irrelevant events} drawn from
fixed templates (shelf, audit, inventory, display notes). The agent
is later queried with a downstream task whose correct execution
requires recalling the seed-episode constraints across hundreds of
intervening turns.

Algorithm~\ref{alg:longhorizon} details the construction. Stage~1
assembles a chronologically timestamped, interleaved event sequence;
Stage~2 streams it through the memory backend so that
event-boundary-driven summarization and long-term consolidation are
triggered \emph{in the loop}, matching real deployment rather than
post-hoc compression. All randomness is seeded by $\mathrm{SHA256}$
of the case identifier for reproducibility. In the longest
configuration the trace contains more than 270 rounds and over 100k
accumulated prompt token length.

\begin{algorithm}[ht]
\caption{Long-Horizon Context Construction with Online Consolidation}
\label{alg:longhorizon}
\begin{algorithmic}[1]
\Require seed episodes $\mathcal{S}_c$, token budget $B$, time gap $\Delta t$,
         templates $\mathcal{T}$, base length $L_0$
\Statex \textbf{// Stage 1: Interleaved sequence assembly}
\State $\mathcal{E} \gets \mathcal{S}_c$;\quad
       $t \gets \max_{e \in \mathcal{S}_c} e.\tau$;\quad
       $\alpha_c \sim \mathcal{U}\left(0.6, 1.6\right)$
\While{$\textsc{tokens}\left(\mathcal{E}\right) < B$}
    \State $T \sim \mathcal{T}$;\quad
           $\ell \sim \mathcal{U}\left(0.5, 1.5\right)\cdot \alpha_c L_0$;\quad
    \State $e \gets \left(\textsc{Sample}\left(T, \ell\right),\ t,\ \texttt{irrelevant}\right)$
    \State $\mathcal{E} \gets \mathcal{E} \cup \{e\}$;\quad
           $t \gets t + \Delta t$
\EndWhile
\State $\mathcal{E} \gets \textsc{SortByTime}\left(\mathcal{E}\right)$
\Statex \textbf{// Stage 2: Online memory consolidation}
\State $\mathcal{M} \gets \emptyset$
\For{$e \in \mathcal{E}$}
    \State $\mathcal{M} \gets \textsc{MemoryBackend}\left(\mathcal{M},\ e\right)$
\EndFor
\State \Return $\mathcal{M}$
\end{algorithmic}
\end{algorithm}

\section{Visual Monitor: Design and Empirical Evaluation}
\label{sec:visual-monitor}

Real-world task completion is evaluated using an independent visual
monitor. After each executed physical action, a separate VLM client
is queried under an empty-context setting and receives only the
chronologically ordered sequence of images from the transient memory
buffer (before, during, and after execution). It is asked to describe
visible changes between successive observations and to determine
whether the resulting physical state supports the next planning step.
This design intentionally decouples task verification from the
planner's reasoning process: since the monitor has no access to the
intended command, task history, or prior chain of thought, its
judgment is grounded in visual evidence rather than inferred intent,
which reduces hallucination-induced verification errors and mitigates
task-context leakage. The resulting state update is returned to the
planner and can trigger replanning upon grasp failure, object
removal, unchanged source-object state, or other physical deviations.
The exact prompt is provided in
Appendix~\ref{sec:visual-differencing-prompt}.

To evaluate whether the visual monitor can serve as an
execution-level supervisor rather than a passive visual captioner, we
construct a balanced verification set from real manipulation
episodes: ten successful pick-and-place executions and ten failed
executions, including both natural failures and externally induced
perturbations such as manually removing the object during execution.
Each candidate VLM receives the same before/during/after triplet and
predicts whether the expected physical state has been achieved.
Table~\ref{tab:verifier-latency} reports mean latency and binary
verification accuracy across candidate backbones. Closed-source
Gemini-3.1-Pro and Qwen3.6-Flash deliver competitive F1 but incur
non-trivial API latency that bottlenecks real-time preemption. Among
locally deployable models, Qwen3-VL-30B-A3B matches Gemini-3.1-Pro's
F1 while running an order of magnitude faster on a dual-H100 server,
offering the best accuracy-latency trade-off. We therefore adopt
Qwen3-VL-30B-A3B as the default visual monitor backbone in all
real-world deployments.

\begin{table}[ht]
\centering
\renewcommand{\arraystretch}{1.35}
\small
\caption{Latency and task-completion verification accuracy of
candidate VLM backbones used as the visual monitor. The 20-trial
balanced set comprises 10 successful and 10 failed/perturbed
pick-and-place executions. A trial is counted as \emph{positive} if
the monitor predicts that the expected post-action state has been
achieved. Closed-source models are accessed via remote API;
open-weight models are deployed locally on a dual-H100 server.}
\label{tab:verifier-latency}
\begin{tabular}{lccccccc}
\hline
Model & Deployment & Latency (s) & TP $\uparrow$ & FN $\downarrow$ & FP $\downarrow$ & TN $\uparrow$ & F1 $\uparrow$ \\
\hline
Gemini-3.1-Pro      & Remote API   & 11.67 & 8 & 2 & 0 & 10 & 0.89 \\
Qwen3.6-Flash       & Remote API   & 7.21  & 9 & 1 & 0 & 10 & 0.95 \\
Qwen3-VL-8B         & Local (2$\times$H100) & 0.69  & 6 & 4 & 0 & 10 & 0.75 \\
Qwen3-VL-30B-A3B    & Local (2$\times$H100) & 0.47  & 8 & 2 & 0 & 10 & 0.89 \\
\hline
\end{tabular}
\end{table}
\section{Planner Prompts}

The semantic planner is prompted to produce exactly one structured next action
at each iteration. The execution history is treated as attempted commands rather
than guaranteed state changes, and post-action visual differencing feedback is
used to update the world state before choosing the next step. The planner must
return \texttt{next\_step: null} only when the task is complete.

\begin{tcblisting}{omniPrompt,title={UR5e Manipulation Planner Prompt}}
# UR5e Reading Desk Planner

You are the user-facing UR5e arm planner for a reading desk setup experiment.
You plan one action at a time from visual evidence and structured runtime
feedback.

Operating loop:
1. At task start, you may receive one initial scene summary for grounding.
2. On every planning step, you receive the current observation image.
3. After each executed action, you may receive post-action visual delta feedback
   comparing before/process/after images.
4. Treat execution history as attempted commands, not guaranteed state changes.
5. Update the world state from the current image and visual delta feedback
   before choosing the next action.
6. Return exactly one next step, or next_step:null when the task is complete.

Allowed actions:
- speak(message)
- store_memory(content, scope, category)
- control_light(action, device, brightness)
- play_audio(action, audio_type, track, volume)
- web_search(url, query)
- set_home_mode(device, option_id, option_text)
- pick_and_place(item_name, source, target)
- get_observation({})

Visual feedback rules:
- Use post-action visual delta feedback to infer what actually changed.
- If visual feedback contradicts requested action parameters, trust observed
  visual changes.
- If an object remains visible at the source after a manipulation attempt, it is
  still pending unless visual delta says otherwise.
- If the target visibly contains the moved object, treat that object as handled.

Return exactly one JSON object:
{
  "current_step_analysis": {
    "visual_state": "Brief current visual state",
    "task_progress": "What has been completed and what remains",
    "next_action_reasoning": "Why the next action follows from the image and feedback"
  },
  "next_step": {
    "step_number": 1,
    "agent": "ur5e_arm",
    "location": "home",
    "action": "pick_and_place",
    "action_type": "act",
    "parameters": {"item_name": "water", "source": "tray", "target": "basket"}
  },
  "needs_human_input": false,
  "user_question": null
}
\end{tcblisting}

\begin{tcblisting}{omniPrompt,title={Keenon Navigation Planner Prompt}}
# Vision-Based Autonomous Humanoid Robot Task Planner

You are the VLM planner for a Keenon robot in a home-navigation benchmark. You
operate in half-open-loop mode: plan exactly one next action, wait for
execution, then plan again from the updated context.

Allowed actions:
- speak(message)
- send_agent_message(message, recipient)
- store_memory(content, scope, category)
- control_air_conditioner(action, temperature, fan_speed)
- control_light(device, action, brightness)
- set_home_mode(device, option_id, option_text)
- play_audio(audio_type)
- web_search(url, query)
- query_weather_api(location, query)
- navigate_to(target_location, with_item="none")
- get_observation(target)

Return exactly one JSON object:
{
  "current_step_analysis": {
    "visual_state": "What is visible, without using it to skip required navigation",
    "task_progress": "What required actions have already been completed",
    "next_action_reasoning": "Why this single next action is required"
  },
  "next_step": {
    "step_number": 1,
    "agent": "Keenon humanoid_robot",
    "location": "current tracked location or unknown",
    "action": "specific_action_name",
    "action_type": "talk|tool|act|sense",
    "parameters": {}
  },
  "needs_human_input": false,
  "human_question": null
}
\end{tcblisting}

\section{Memory Compression Prompts}

\begin{tcblisting}{omniPrompt,title={Episodic Memory Compression Prompt}}
System prompt:
You are an episodic interaction log writer for an embodied planner. Summarize
what happened in this interaction segment as an operational log, not as
long-term memory. Preserve chronology, task context, concrete observations,
tools, skills, actions, useful parameters, outcomes, errors, and unresolved
questions. Do not infer durable user preferences unless the user explicitly
stated one; if present, keep it as evidence rather than as the main focus.
Return strict JSON with keys: topic, timestamp, source_turn_range, task_request,
interaction_summary, actions_or_skills, outcome, important_observations,
open_questions, evidence_snippets.

User prompt template:
Level: Episodic
Device: {profile_name}
Trigger: {trigger_reason}
Suggested topic: {topic}
Summary timestamp: {created_at}
Source turn range: {source_turn_range}

Write this as an episodic interaction log. Prioritize timestamped task flow,
user request, planner actions or skills, execution result, and concrete
feedback. Do not compress it into stable preference rules; long-term summaries
will handle cross-session preference and experience extraction.

Source text:
{source_text}
\end{tcblisting}

\begin{tcblisting}{omniPrompt,title={Long-Term Memory Compression Prompt}}
System prompt:
You are a memory compaction assistant for an embodied planner. Do not summarize
chronology. Extract the durable decision rules future planning must obey. Return
strict JSON with keys: topic, hard_preferences, soft_preferences, current_needs,
time_constraints, forbidden_items, decision_rules, stable_user_preferences,
priority_order, evidence_snippets.

Guidance:
Extract only the information that should affect future decisions. Prefer durable
user preferences and hard constraints over one-off event narration. If a current
temporary need exists, keep it in current_needs instead of
stable_user_preferences. For long-term summaries, read episodic interaction logs
as evidence and extract stable user preferences, recurring constraints, and
reusable task execution experience. Use decision_rules for reusable planning or
skill-selection lessons learned from repeated interactions.

User prompt template:
Level: long_term
Device: {profile_name}
Trigger: {trigger_reason}
Suggested topic: {topic}

{level_specific_guidance}

Source text:
{source_text}
\end{tcblisting}

\section{Visual Differencing Prompt}
\label{sec:visual-differencing-prompt}

\begin{tcblisting}{omniPrompt,title={Visual Differencing Prompt}}
You are an independent visual differencing model for a robot planner.

You are not given the robot's intended command. Do not infer what should have
happened. If an object appears unchanged, say it remains unchanged. If the
images are ambiguous, say so.

Compare the images in temporal order. Focus on visible changes between the
images. Try to enumerate all scene changes that could matter to the next robot
planning step, including:
- object presence, absence, position, orientation, and containment changes
- objects that remained in their original place despite nearby motion
- gripper or arm pose changes, if visible
- occlusions, blur, lighting/camera viewpoint shifts, or other uncertainty sources

Do not describe the whole scene from scratch. Do not assume an action succeeded.
Do not use prior task context. Prefer concrete visual evidence over short
generic summaries. If there is no meaningful visible change, say that explicitly.

Return JSON only with this schema:
{
  "observed_changes": ["detailed visible changes, one fact per item"],
  "objects_moved_or_removed": [
    {
      "object": "name or description",
      "from": "observed source",
      "to": "observed destination or unknown",
      "evidence": "specific visual evidence"
    }
  ],
  "objects_remaining": [
    "relevant objects still visible, including their observed locations"
  ],
  "gripper_or_arm_state": "visible state and pose change if relevant, otherwise unknown",
  "scene_change_notes": [
    "camera, lighting, occlusion, blur, or ambiguity notes"
  ],
  "action_context_used": false,
  "contradicts_requested_action": false,
  "state_update": "planner-facing summary covering the key visible changes and unchanged relevant objects",
  "confidence": "low|medium|high"
}
\end{tcblisting}

\end{document}